\def\year{2021}\relax
\documentclass[letterpaper]{article} 
\usepackage{aaai21}  
\usepackage{times}  
\usepackage{helvet} 
\usepackage{courier}  
\usepackage[hyphens]{url}  
\usepackage{graphicx} 
\usepackage{natbib}  
\usepackage{caption} 
\title{The EpiBench Platform to Propel AI/ML-based Epidemic Forecasting: \\A Prototype Demonstration Reaching Human Expert-level Performance}
\author{
    Ajitesh Srivastava,
    Tianjian Xu,
    Viktor K. Prasanna
    \\
}
\affiliations{

    University of Southern California\\
    Los Angeles, USA
    \{ajiteshs, frostxu, prasanna\}@usc.edu

}

\begin{document}

\maketitle

\begin{abstract}
During the COVID-19 pandemic, a significant effort has gone into developing ML-driven epidemic forecasting techniques. However, benchmarks do not exist to claim if a new AI/ML technique is better than the existing ones.
The ``covid-forecast-hub'' is a collection of more than 30 teams, including us, that submit their forecasts weekly to the CDC. It is not possible to declare whether one method is better than the other using those forecasts because each team’s submission may correspond to different techniques over the period and involve human interventions as the teams are continuously changing/tuning their approach. Such forecasts may be considered ``human-expert'' forecasts and do not qualify as AI/ML approaches, although they can be used as an indicator of human expert performance. We are interested in supporting AI/ML research in epidemic forecasting which can lead to scalable forecasting without human intervention. Which modeling technique, learning strategy, and data pre-processing technique work well for epidemic forecasting is still an open problem. To help advance the state-of-the-art in AI/ML applied to epidemiology, a benchmark with a collection of performance points is needed and the current ``state-of-the-art'' techniques need to be identified. We propose EpiBench a platform consisting of community-driven benchmarks for AI/ML applied to epidemic forecasting to standardize the challenge with uniform evaluation protocol. In this paper, we introduce a prototype of EpiBench which is currently running and accepting submissions for the task of forecasting COVID-19 cases and deaths in the US states and  We demonstrate that we can utilize the prototype to develop an ensemble relying on fully automated epidemic forecasts (no human intervention) that reaches human-expert level ensemble currently being used by the CDC.

\end{abstract}

\section{Introduction}

\begin{figure*}
    \centering
    \includegraphics[width=0.6\textwidth]{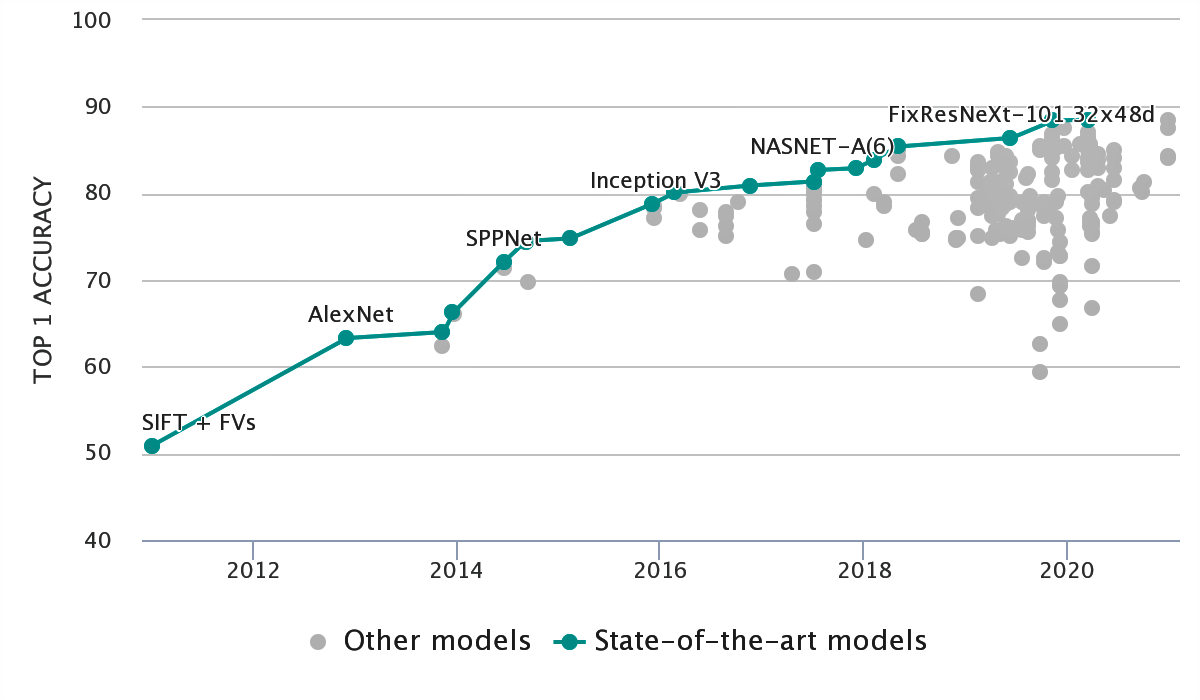}
    \caption{Increasing interest in image classification over time with best performances over the years on ImageNet~\cite{papers-with-code}. Benchmarks like ImageNet helped propel Computer Vision research.}
    \label{fig:imagent}
\end{figure*}


As of writing this paper, a Google Scholar search for ``covid forecasting'' yields 14000 results. AI/ML conferences are seeing an increasing number of papers on epidemic forecasting, all of which lack a well-defined benchmark. While a researcher may compare their approach against a traditional modeling strategy such as SEIR~\cite{hethcote2000mathematics}, our experience suggests that the forecasts are highly sensitive to the implementation which requires deciding data pre-processing and learning strategy. Consequently, ``SEIR modeling'' is only a part of the forecasting process, but itself cannot be considered a forecasting method to compare against; it only describes the epidemic process, but does not dictate how to pre-process the data and learn the model parameters, which are critical and non-trivial decisions. This is analogous to how one cannot simply compare against ``deep learning'' and must declare the structure of the network and the learning algorithm.
Therefore, in the current state, performing an evaluation and claiming whether an AI/ML new approach is better or when one approach performs better than the other is not possible. There is a need to define the boundary by recognizing the state-of-the-art approaches so that new research can be conducted and verified to push the boundary. 

The existence of such benchmarks has propelled other fields in AI/ML. For instance, datasets like ImageNet~\cite{deng2009imagenet} and CIFAR-100~\cite{krizhevsky2009learning} along with methods like ResNet~\cite{he2016deep} and VGG-16~\cite{simonyan2014very} have created a standard against which new approaches need to be compared for Image Classification. Since the release of ImageNet in 2013, the accuracy has been improved from $51\%$ to almost $90\%$ as seen in Figure~\ref{fig:imagent}. The Figure also shows an increasing interest in the task on that dataset over the years as an increasing number of ``other models''. Standard benchmarks have also propelled research in other fields such as Speech~\cite{speech-benchmark} and Graph Neural Networks~\cite{ogbn-benchmark}.

We propose EpiBench a platform to bring the same benefits to AI/ML applied to epidemic forecasting. We are creating a platform for the community to engage in discussions regarding identification of critical modeling and learning decisions, a community-driven ensemble approach development, and  identification of proper evaluation techniques. The platform will be easily extensible by the community in terms of adding new forecasting tasks and datasets. Finally, while EpiBench focuses on retrospective forecasting, i.e., forecasting when ground truth is already available, the approaches identified to be the state-of-the-art will benefit real-time forecasting efforts like Epidemic Prediction Initiative~\cite{EPI} by the CDC, and other agencies around the world which drive the Government's response. In this paper, we make the following contributions:

\begin{itemize}
    \item We introduce a prototype of EpiBench which is currently running and accepting submissions for the task of forecasting COVID-19 cases and deaths in the US states and counties.
    \item We demonstrate that we can utilize the prototype to develop an ensemble relying on fully automated epidemic forecasts (no human intervention) that reaches human-expert level ensemble currently being used by the CDC.
    \item We demonstrate how EpiBench can identify critical decisions in AI/ML-based epidemic forecasting.
\end{itemize}
We emphasize that this paper is not intended to demonstrate a fully-fledged benchmarking platform, nor to show that a new forecasting method is superior. Instead, the intention is to demonstrate the need for such a platform and results from a prototype as stated above in the contributions, that indicate the potential of the envisioned platform.


\section{Background}
\subsection{Existing Infrastructure}
We first describe some recent platforms that have emerged for submissions and evaluations of COVID-19 forecasts and draw a contrast from our goal.
\subsubsection{COVID-19 Forecast Hub}
The COVID-19 forecasting hub~\cite{ray2020ensemble} keeps a ``live" record for forecasts of COVID-19 in the US, created by more than 30 leading infectious disease modeling teams from around the globe, including ours, in coordination with CDC. Every week the teams submit their latest forecasts for the number of cumulative and/or incident cases and deaths for US states, counties, and at the national level. The forecasts are reported at a weekly granularity for the upcoming Sundays. This ensures that the periodicity in reporting does not affect the evaluation. Furthermore, daily forecasts are not being used or being communicated~\cite{ray2020ensemble}.

\begin{figure*}[!ht]
    \centering
    \includegraphics[width=\textwidth]{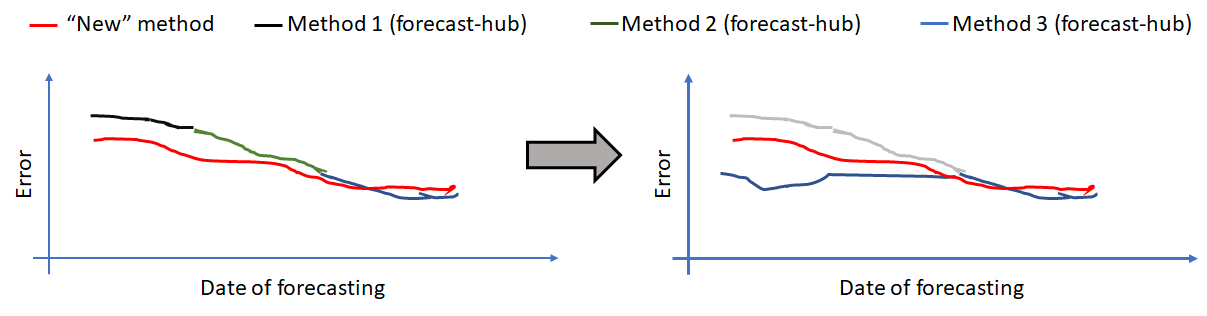}
    \caption{Comparison of a new methodology against the forecast hub may be misleading as the submitted forecasts may be a combination of various methods.}
    \label{fig:fh-comparison}
\end{figure*}

The submitted forecasts may be considered real-time expert forecasts, although, they may not qualify as an AI/ML methodology, although parts of forecast generation may use AI/ML. First, the forecasts represent the submissions by a team and not necessarily by a particular method -- the teams are allowed to change their methods. While this platform is useful to provide the best guess of the experts in real-time (as an epidemic is unfolding), due to varying methodologies over different periods of time, it does not provide any insight into which methods work and when. A comparison of a new methodology against the existing submissions will lead to misleading conclusions. For instance, suppose a new methodology computes its errors at different points in time and compares against the errors of a particular team's submission from the forecast hub (see Figure~\ref{fig:fh-comparison}). From this comparison, one may falsely conclude that the new methodology is better. However, the submitted forecasts may be produced by three different methodologies over time, one of which when used for retrospective forecasts may produce better results. Further, many submissions are tuned by the experts for every submission, through trial-and-error, visual inspection, and may derive knowledge from other sources that can evolve over time. Such methods may not classify as an AI/ML method as they require significant human involvement and will not scale to a large number of time-series. Yet, the submitted forecasts provide a collection of baselines that may be used as performance points representing the real-time expert forecasts, without the claim of one method being better than the other.

\subsubsection{Epidemic Prediction Initiative}
While the COVID-19 forecasting hub has received a lot of attention recently, it is a part of the Epidemic Prediction Initiative (EPI) by the CDC~\cite{EPI}.  EPI aims to improve the science and usability of epidemic forecasting by facilitating open forecasting projects and challenges. EPI provides access to some data and defines a metric with which the forecasts are going to be evaluated. EPI has several forecasting challenges including COVID-19, Aedes, West Nile Virus, Dengue, and Influenza. When deciding on the state-of-the-art methodology and comparing a new approach against an existing one, EPI suffers from the same issues as the COVID-19 forecast hub. EpiBench will complement the efforts of EPI and COVID-19 forecast hub-like initiatives by benchmarking methodologies instead of the comparison of teams while utilizing similar data and expertise from the community. The existing initiatives will continue to assist ongoing epidemics through the collective wisdom of the experts, while EpiBench will provide a platform for ``retrospective'' forecasting to compare and learn from and encourage the development of new AI/ML methods.

\subsubsection{Existing Evaluations for COVID-19}

Particularly, for COVID-19, some teams who regularly submit their forecasts to the CDC have been performing periodic evaluations of all the submissions. For example, 
Youyang Gu evaluates week-ahead predictions for cases and death for county, state, and national level forecasts for the US\footnote{\url{https://github.com/youyanggu/covid19-forecast-hub-evaluation}} using mean absolute error~\cite{willmott2005advantages}. The team Caltech-CS156\footnote{http://cs156.caltech.edu/model/compare.html} evaluate daily death forecasts over 7-day period and weekly forecasts.
Steve McConnell\footnote{\url{https://github.com/stevemcconnell/covid19-forecast-evaluations}} performs a detailed evaluation of death forecasts of all the submitted models using a number of metrics that address the drawbacks of the traditional evaluation metrics.
A forecast hub for Germany and Poland also evaluates submitted forecasts\footnote{\url{https://jobrac.shinyapps.io/app\_evaluation/}} for country-level cases and deaths, although they seek submissions of Admin1-level (state) as well. A detailed comparison of publicly available death forecasts for countries around the world is tracked by Covidcompare\footnote{http://covidcompare.io}.
Our own evaluations are available on our webpage~\footnote{\url{https://scc-usc.github.io/ReCOVER-COVID-19/\#/leaderboard}}, where we present mean absolute errors over time for all the methods submitted to the CDC for US state-level case and death forecasts, and for those submitted to Poland+Germany forecast hub for country and admin1-level case and death forecasts.

The fundamental difference between these evaluations and our goals is that they compare the performance of forecasters rather than AI/ML methods. This is because (i) not all forecasts are generated using AI/ML approaches, (ii) forecasts may be manually tweaked before submission by a human-expert, and (iii) over time forecasters may keep changing their methodology. While such evaluations are crucial for understanding where each team stands every week as the epidemic unfolds, it provides us with no insight on what methods work well and thus does not add to our understanding of how epidemic forecasting can be improved without human intervention.


\subsection{Acceptable Methods}

EpiBench aims to enable AI/ML research in epidemic forecasting that currently has a big hurdle of lack of proper benchmarks. It will create a platform for sustained research in improving AI/ML methods as well as our understanding of which technique works when. Before we proceed, we clarify what qualifies as AI/ML in this context.

We define an \textbf{AI/ML approach} or \textbf{methodology} as \textit{a technique that is a self-contained algorithm/code that does not rely on human intervention for generating different forecasts within the same task and has no foresight, i.e., does not use information/data that was available after the date at which the forecast is being generated.} For instance, suppose, the task is to forecast the number of deaths in the US states due to COVID-19 in the next 5 weeks of every Sunday in the period of April to October. Let $\mathcal{M}$ be a fixed function determined by the AI/ML methodology. Then for any date $t_i$ in the period, with the version of historical data $D_i$ as it was seen on $t_i$, the generated forecasts should be $\mathcal{M}(D_i)$, without any additional inputs. Any change in the methodology, including simple data pre-processing change, is considered to lead to a different methodology. Further, $\mathcal{M}$ should not have any direct dependence on the $t_i$. This excludes explicit programming of using one methodology for an interval and a different one for another. Any approach that does not qualify based on the above but does not use any foresight will be considered \textbf{human expert forecasts}. While the state-of-the-art will be defined by AI/ML methods in EpiBench, human expert forecasts will also be included to indicate human expert performance. Submissions in COVID-19 forecast-hub and EPI will be considered under this category. 

We focus on advancing AI/ML approaches as the immediate target for EpiBench because AI/ML approaches are more scalable. There are more than 3000 counties in the US for which the data is readily available. 
Google COVID-19 Open Data Initiative\footnote{\url{https://github.com/GoogleCloudPlatform/covid-19-open-data}} provides epidemiological data for around 20,000 locations.
Due to the global crisis experienced worldwide during COVID-19, we envision that more data will be available for the future epidemics, resulting in a large number of time-series infeasible to be manually analyzed. Therefore, there is a need for AI/ML-driven fully automated approaches and a platform for proper evaluation and to set up the community for the following innovations, which will be enabled by EpiBench.

\section{Results from the Prototype}
We have developed an early prototype of EpiBench called COVID-19 forecast-bench~\cite{forecast-bench} for the task of  COVID-19 case and death predictions in the US states and counties. It keeps a daily record of the versions of case and death time-series as reported by JHU CSSE COVID-19 Data~\cite{JHU}. The submitting teams are requested to provide details of their methodologies, particularly regarding data pre-processing, modeling technique, and learning strategy. It is clarified that even changing a simple decision like the smoothing window qualifies as a different methodology. We have already received 3 AI/ML methods excluding several of our own. We have also included more than 30 methodologies (human expert-driven) pulled from the COVID19 forecast-hub~\cite{forecast-hub}. Currently, we are providing evaluations for 1-, 2-, 3-, and 4-week ahead forecasts using mean absolute error~\cite{willmott2005advantages}:
For each prediction $\hat{y}_i$ corresponding to the ground truth $y_i$, $i \in \{1, \dots, n\}$, mean absolute error (MAE) is defined as:
\begin{equation}
    MAE = \sum_i \frac{|y_i - \hat{y}_i|}{n}\,.
\end{equation}
For the prototype, we define the ground truth as the positive cases and deaths timeseries reported in the latest version of JHU CSSE COVID-19 Data. It is known that this data changes over time, due to back correction~\cite{JHU}. Therefore, the latest version of the data is the best known approximation of the true targets (positive cases and dates) for a prior date. 

\subsection{Ensemble Development}

\begin{figure*}
    \centering
    \includegraphics[width=0.6\textwidth]{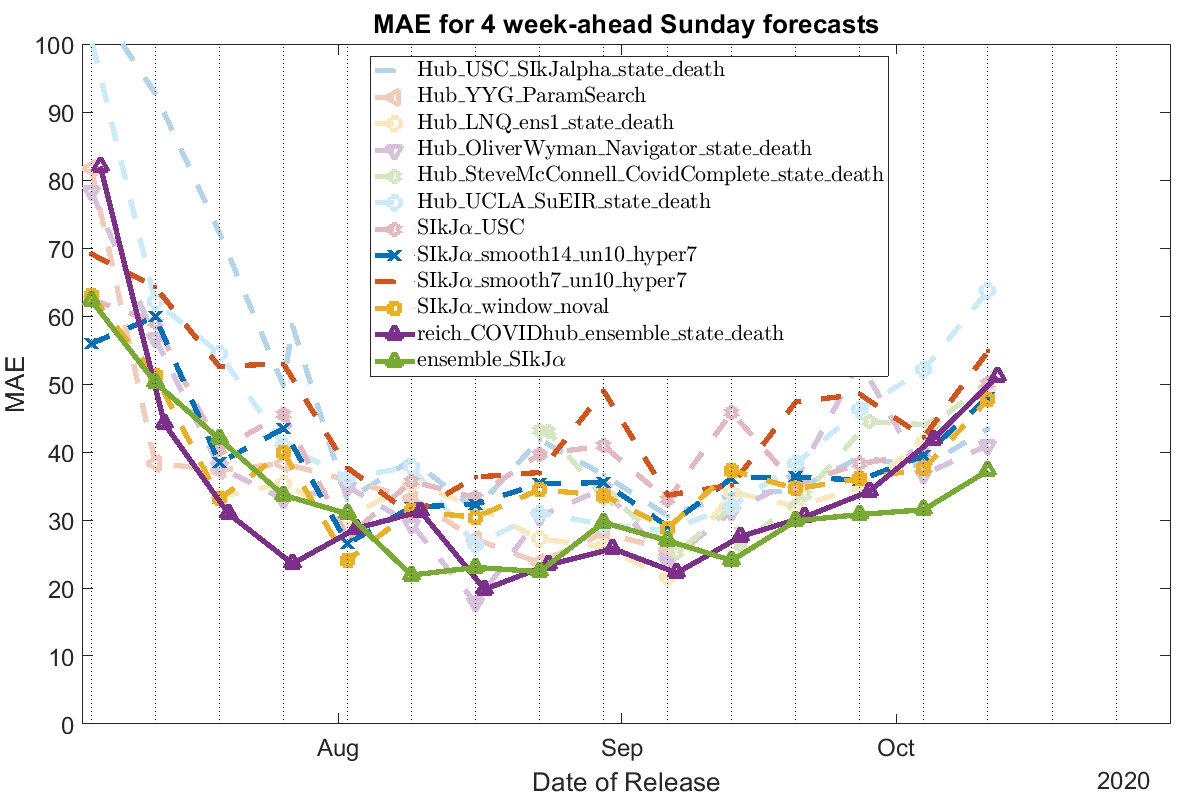}
    \caption{Comparison of the ensemble approach built on EpiBench AI/ML-based approaches against human expert forecasts and their ensemble.}
    \label{fig:ensembleRF}
\end{figure*}

One advantage of having retrospective forecasts with various methodologies is to be able to design intelligent ensemble techniques. On the other hand, the ensemble approach of the forecast-hub is limited to taking the average of individual submissions~\cite{ray2020ensemble}. This is due to the fact that the submissions by individual teams do not stick to the same methodology and involve human decision making. Therefore, learning a model treating the past submissions as predictors is not useful as the nature of the predictors keeps changing. 

\subsubsection{Constituent Methodologies as ``Predictors''}
To demonstrate the utility of retrospective forecasts in ensemble development, we use variations of our SIkJ$\alpha$ approach~\cite{srivastava2020fast}. This approach models the epidemic as a discrete-time process with temporally varying infection and death rates.
The model considers many complexities such as unreported cases due to any reason (asymptomatic, mild symptoms, willingness to get tested), immunity (if any) or complete isolation, and reporting delay, and yet, it can be reduced to a system of two linear equations which can be fitted one
after the other resulting in fast yet reliable forecasts. The forecasts are performed by smoothing the data, learning the parameters by fitting the reduced model, and then simulating the model into the future using these learned parameters.

We use three variations of our SIkJ$\alpha$ approach for the task of predicting deaths in the US states over the period of July to October:
\begin{itemize}
    \item SIkJ$\alpha$\_smooth14\_un10\_hyper7: This indicates that the input data was set to smooth over 14 days, the under-reporting factor was set to 10 (this input has little effect on short-term forecasts~\cite{srivastava2020fast}), and the hyper-parameters were learned on a held-out validation data of the last 7 days.
    
    \item SIkJ$\alpha$\_smooth7\_un10\_hyper7: This indicates that the input data was set to smooth over 7 days.
    
    \item SIkJ$\alpha$\_window\_noval: This indicates that the input data was set to smooth over 14 days and the hyper-parameters were learned by fitting on a window of the past few days (itself treated as a hyper-parameters) without a separate validation set.
\end{itemize}
Each of the above has advantages and disadvantages. SIkJ$\alpha$\_smooth14\_un10\_hyper7 is able to avoid noise by smoothing over a large window, while it may over-smooth a newly emerging pattern which is better detected by SIkJ$\alpha$\_smooth7\_un10\_hyper7. Having a  small validation set as in these two approaches helps identify hyper-parameters during changing trends. However, this may also overfit, and during stable trends picking the hyper-parameters based on fitting over a large window could be better.
Note that our SIkJ$\alpha$ is fast ($\approx4$s for all US states per run), and so we are able to quickly generate different sensible variations. Next, we will use the results of these methodologies as the constituent predictors in the ensemble.

\subsubsection{Ensemble using Random Forests}

We wish to utilize the individual predictors to improve the forecasts. We generate the forecasts using each of the above on each Sunday using the version of the data that was available on that day, for a period of July-Oct and use them as input predictors in a Random Forest~\cite{meinshausen2006quantile}. Additionally, it takes the cumulative and incident deaths that week and the incident deaths on the previous week as inputs. The objective is to find a regression model that can consider the outputs of the above three approaches and the recent trend, and ``readjust'' them to reduce error in the future. The Random Forest was implemented using the Matlab TreeBagger function with 100 trees\footnote{https://www.mathworks.com/help/stats/treebagger.html}. For each Sunday, the Random Forest was trained using the outcomes in the last two weeks for which ground truth is available. For 4-week ahead forecast, this would mean training on the forecasts generated 4-5 weeks ago by the three approaches listed above. The code to reproduce our results is available in the forecast-bench repository~\footnote{https://github.com/scc-usc/covid19-forecast-bench}.

Figure~\ref{fig:ensembleRF} shows the errors obtained on 4-week ahead predictions as a function of the date of releasing the forecast. The ground truth is the latest version of the death time-series available from JHU CSSE~\cite{JHU}.
We compared our ensemble approach to the following:
\begin{itemize}
    \item The individual constituent AI/ML approaches: These are shown in dark dashed lines and include SIkJ$\alpha$\_smooth14\_un10\_hyper7, SIkJ$\alpha$\_smooth7\_un10\_hyper7, and SIkJ$\alpha$\_window\_noval.
    
    \item Top human-expert forecasts submitted to the CDC: These are shown in blurred dashed lines and include submissions from YYG Paramsearch, LNQ ensemble, Oliver Wyman Pandemic Navigator, Steve McConnell CovidComplete, UCLA SuEIR, and our own submission USC SIkJalpha, all of which are available on the forecast-hub. We have also included a version of our model that appears on our Github repository\footnote{https://github.com/scc-usc/ReCOVER-COVID-19} and may differ slightly from our submissions to the forecast-hub.
    \item COVID Hub Ensemble: This is the ensemble (average) of all the submitted forecasts shown in solid dark line. This has been shown to perform better or on par with individual submissions~\cite{ray2020ensemble}.
\end{itemize}

We observe that our ensemble (dark solid purple line) results in a lower error compared to the constituent AI/ML approaches (dark dashed lines). It has a competitive performance with the forecast-hub ensemble and outperforms it consistently over the last few weeks. We emphasize that our ensemble has no human intervention and the whole process \textit{from loading the data to producing the forecasts is fully automated}. This demonstrates that our ensemble can utilize AI/ML approaches to achieve human-expert level performance. We envision that the ensemble error will further reduce through the inclusion of more AI/ML-driven forecasts. 

It is to be noted that we do not claim that our ensemble approach is the best AI/ML-based epidemic forecasting approach because such a claim will require more AI/ML methodologies to be included. This is precisely where EpiBench will make an impact by providing a platform for new methodologies to compare against existing ones under a uniform evaluation protocol.

\subsection{Assessing Forecasting Decisions}
\label{sec:inno-forecasting}
Going from data to forecasts requires many decisions that can influence the accuracy. These include, but not limited to, the following:
\begin{itemize}
    \item Data pre-processing: Whether to smooth the data over 7 days or 14 days, picking a rule to declare a point anomalous and a rule to replace anomalous data points.
    \item Choice of modeling: Whether to use an SEIR model, a deep learning model, or an ARIMA model.
    \item Learning strategy: Whether to use Bayesian learning, least square regression, or gradient descent.
    \item Hyper-parameter setting: Specific structure of the neural network, size of the window to consider, or a forgetting factor/weighted least square to only account for the latest trends.
\end{itemize}

\begin{figure}
\centering
  \centering
  \includegraphics[width=\columnwidth]{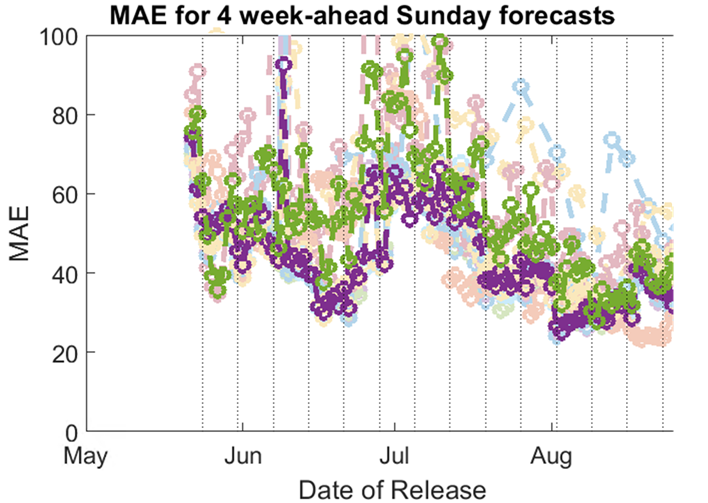}
  \captionof{figure}{The choice of smoothing window can significantly affect the error.}
  \label{fig:smooth-diff}
\end{figure}

 Each of the above decisions may lead to different outcomes. Despite the plethora of work in forecasting this year, it is not known which of the above decisions are more critical. As an example, consider the two dark lines (green and purple) in Figure~\ref{fig:smooth-diff} plotted against the blurred lines corresponding to the performance of submissions at the COVID-19 forecast-hub. The two dark lines (green and purple) show a significant difference in their performance, yet the two methodologies are identical with the exception that one of them smooths the data over 14 days and the other over 7 days. Many other approaches perform in the middle of the two while being very different methodology-wise as they use deep learning or Bayesian learning as opposed to our regression approach. This may indicate that for this task, the outcome is more dependent on data pre-processing rather than the choice of the learning strategy. Such an observation can accelerate the computation as regression can be performed much faster than Bayesian learning, and also, it can direct the research in improving data pre-processing and noise filtering which may be more critical for this task and similar tasks of the future.

\begin{figure}
  \centering
  \includegraphics[width=\columnwidth]{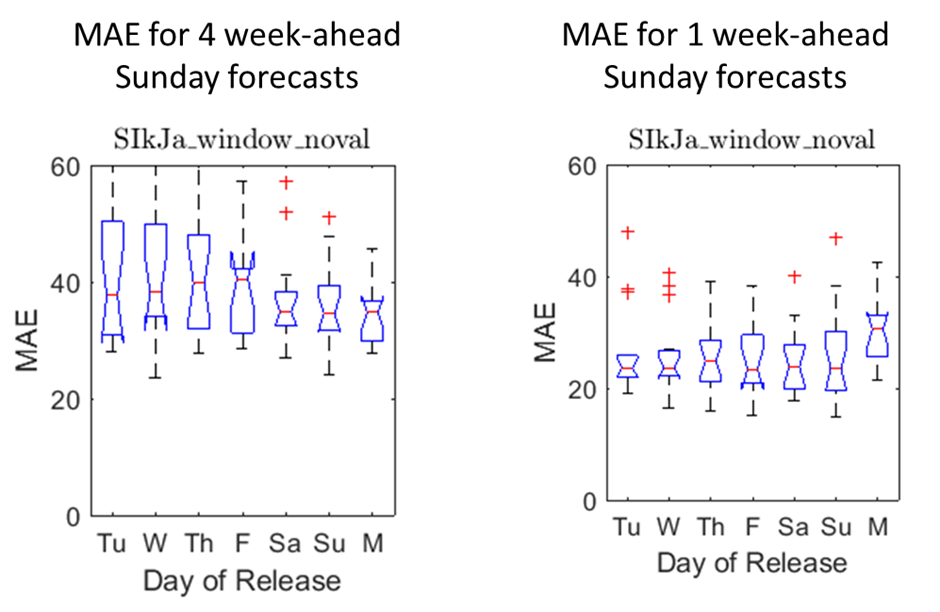}
  \captionof{figure}{The choice of the day of forecast can significantly affect the error.}
  \label{fig:day-diff}
\end{figure}

  Figure~\ref{fig:day-diff} shows the distribution of errors obtained by a method over different days of the week. The plot indicates that there can be a significant difference in performance based on when the forecasts are performed. Again, this suggests that there may be some simple decisions, often overlooked, that affect the outcome. While we have performed these analyses only on variations of our own model SIkJ$\alpha$, it is likely to be the case for other models as well. This is supported by the fact that many of these submitted forecasts claim to use SEIR model\footnote{\url{https://www.cdc.gov/coronavirus/2019-ncov/covid-data/forecasting-us.html}}, yet they produce drastically different outcomes. This reinforces our claim that decisions beyond the choice of a model are critical in producing accurate forecasts.
  EpiBench will enable us along with the research community in identifying which decisions are critical by presenting the evaluation of many different methodologies and identifying them based on specific decisions they make to arrive at their forecasts.


\section{Discussions}
\begin{figure*}[!ht]
    \centering
    \includegraphics[width=\textwidth]{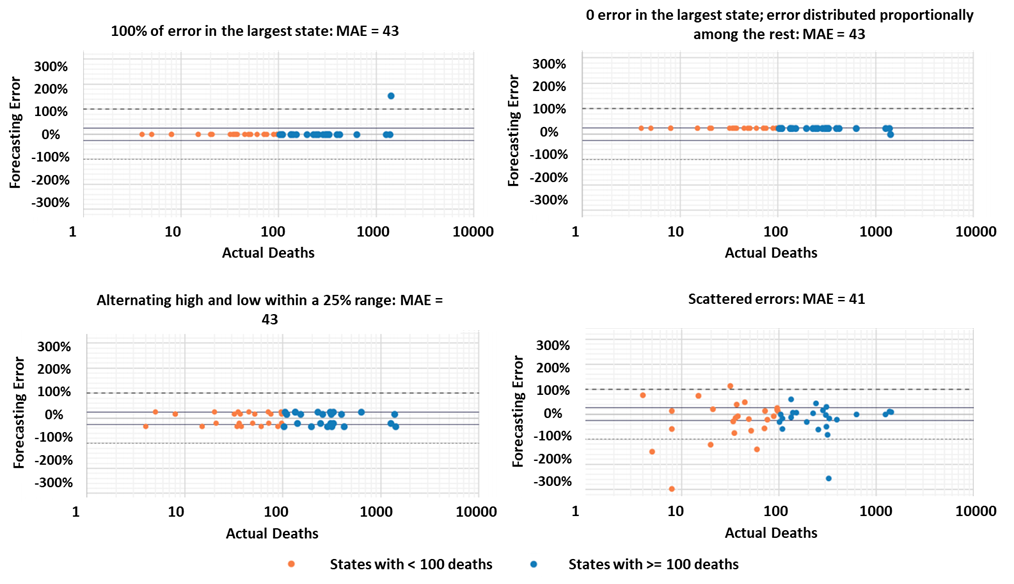}
    \caption{Drawback of Mean Absolute Error (MAE): May assign a low error to a result that intuitively is worse. (These plots are \textcopyright~ Copyright Steve McConnell. Additional information is available at~\cite{steve-covid})}
    \label{fig:MAEdrawback}
\end{figure*}
\subsection{Evaluation Protocol}
We wish to leverage EpiBench to start a discussion that will expose the advantages and drawbacks of various metrics and identify a set of metrics for point, interval, and probabilistic forecasts to provide a uniform evaluation protocol. We are planning to initially use the following well-known error metrics, categorized with respect to the type of forecast output.

\textbf{Point forecasts} refer to the forecasting of one number for each ground truth of the future, such as predicted number of new cases on a future date and declaring a future date when the peak is expected to happen. For such forecasts we will use mean absolute error (MAE) defined for each prediction $\hat{y}_i$ corresponding to the ground truth $y_i$, $i \in \{1, \dots, n\}$ and a variation of mean percentage absolute error called symmetric MAPE or SMAPE~\cite{o1998judgmental}:
\begin{equation}
    MAE = \sum_i \frac{|y_i - \hat{y}_i|}{n}\,.
\end{equation}
\begin{equation}
    SPAME = \frac{1}{n}\sum_i \frac{|y_i - \hat{y}_i|}{0.5|y_i + \hat{y}_i|}\,.
\end{equation}

Variations of both MAE and MAPE are widely used in time-series forecasting~\cite{chatfield2000time}. Particularly, MAE is the evaluation preferred by the CDC for point forecasts~\cite{EPI}.

\textbf{Probabilistic forecasts} refer to the forecasting of a probability distribution for ground truth of the future, i.e., assigning a probability to each discrete possibility. 
For such forecasts, we will initially use a log score~\cite{gneiting2007strictly} which is used by the CDC to evaluate real-time submissions of Flu forecasts~\cite{EPI}.
\begin{equation}
    LS= \frac{1}{n}\sum_i \max \{\ln{P(E_i)}, -10\}\,,
\end{equation}
where $P(E_i)$ is the probability assigned to the event $E_i$ that is observed in the ground truth. If the assigned probability is so low that $\ln{P(E_i)}$ is less than $-10$ or undefined, it is replaced by $-10$. This ensures that one significantly poor score does not affect the average. A higher score is preferred.

\textbf{Interval forecasts} refer to the reporting of a range with a confidence interval suggesting the likelihood of the true value falling in the range. One way to evaluate prediction intervals is ``coverage"~\cite{ray2020ensemble} that measures the percentage of time the observed value falls in the provided interval for given confidence (such as 95\% confidence interval). Other ways of evaluating interval forecasts and more generally quantile forecasts while penalizing long ranges also exist in the literature~\cite{gneiting2007strictly}.

While the above-mentioned metrics have been widely used in epidemic forecasting and time-series forecasting, none of them are unanimously accepted to be the best metric, and each one suffers from drawbacks. For instance, aggregating errors using MAE may lead to assigning a low error to a result that intuitively is worse (see Figure~\ref{fig:MAEdrawback}). We will perform a detailed study of drawbacks of the initial metrics, and engage the community in the design and selection of future evaluation metrics.

\subsection{EpiBench: Planned Functionalities}
Expanding from our prototype which is in the form a GitHub repository, We plan to create an infrastructure consisting of a website, a GitHub organization and a Slack workspace. The user will be able to perform the following actions using the EpiBench platform:
\begin{itemize}
    \item View current evaluations for submitted models and a list of state-of-the-art based on various metrics for different epidemic forecasting tasks on the EpiBench website.
    \item Upload a set of forecasts for evaluation through a Git pull request to the appropriate repository in the collection. A link to the proper repository will be available on the EpiBench website.
    \item Download code for evaluation and top among the previously submitted forecasts available on our GitHub. This top subset determines the state-of-the-art based on a selected evaluation metric and helps the download and repository maintenance easier by limiting the size.
    \item Access all the submitted forecasts through a Google Drive link on the website.
    \item Participate in discussions on specific code/repository by opening a ``GitHub Issue". Also, discussion on evaluation metrics, new forecasting tasks will be facilitated through Slack~\cite{slack}.
    \item  Upload code for execution to reproduce results and to run the code on more datasets.
\end{itemize}

\section{Conclusions}
We have demonstrated a prototype of EpiBench, a platform planned to provide a benchmark to compare AI/ML methodologies for epidemic forecasting against each other and against human-expert forecasts. We aim to define the state-of-the-art using EpiBench which will help the AI/ML community to push the boundary. Having various AI/ML methods can also lead to robust and accurate ensemble forecasting. We show that using our own AI/ML approaches (fully automated, no human intervention) in the prototype, we are able to develop an ensemble that matches an ensemble of human expert-level performance on COVID-19 death forecasting in the US states. Moreover, EpiBench can help us identify which decisions (data pre-processing, modeling choice, learning strategy, hyper-parameter tuning, etc.) in the forecasting approach are critical in epidemic forecasting and help direct the research accordingly. In the prototype, we have shown that the forecasts may be highly sensitive to data pre-processing. This relates back to the ensemble development as we were able to utilize the forecasts from different data pre-processing (smoothing) and hyper-parameter tuning methods to obtain significant improvements in the ensemble without altering the choice of modeling or learning strategy.

In the future, we wish to expand EpiBench to various epidemic forecasting tasks. The platform will be available to the AI/ML researchers and epidemiologist through (i) the EpiBench website for viewing current evaluations of AI/ML approaches and a list of state-of-the-art based on various metrics for different epidemic forecasting tasks; (ii) a GitHub organization with repositories for uploading forecasts for evaluation and downloading evaluation code and previously submitted forecasts; (iii) a Slack workspace for community discussion on evaluation, new forecasting tasks, and dissemination of updates.

\section{Acknowledgments}
This work was supported by National Science Foundation Award No. 2027007.
We would like to thank Youyang Gu, Dean Karlen and UMich-RidgeTfReg team from the University of Michigan for their initial submissions to the forecast-bench. We also thank Steve McConnell for useful discussions on evaluation metrics.

\bibliography{sample}

\end{document}